\documentclass[dvipsnames,format=sigconf,anonymous=false,review=false,nonacm=true, authorversion=true]{acmart}

\AtBeginDocument{%
  }

\usepackage{subfigure}
\usepackage{float}
\usepackage{dblfloatfix}
\usepackage{placeins}
\usepackage{cuted}

\expandafter\def\expandafter\normalsize\expandafter{%
    \normalsize%
    \setlength\abovedisplayskip{1em}%
    \setlength\belowdisplayskip{1em}%
    \setlength\abovedisplayshortskip{0.5em}%
    \setlength\belowdisplayshortskip{0.5em}%
}

\graphicspath{{img}}

\begin{document}

\title{Neuroevolution of Self-Attention Over Proto-Objects}

\author{Rafael C. Pinto}
%%\authornote{Both authors contributed equally to this research.}
%%\orcid{1234-5678-9012}
\affiliation{%
  \institution{Federal Institute of Education, Science and Technology of Rio Grande do
Sul (IFRS)}
  \city{Canoas}
  %%\state{Rio Grande do Sul}
  \country{Brazil}
}
\email{rafael.pinto@canoas.ifrs.edu.br}

\author{Anderson R. Tavares}
\affiliation{%
  \institution{Federal University of Rio Grande do Sul (UFRGS)}
  \city{Porto Alegre}
  %%\state{Rio Grande do Sul}
  \country{Brazil}
}
\email{artavares@inf.ufrgs.br}

%%\renewcommand{\shortauthors}{Trovato et al.}

\begin{abstract}
Proto-objects --  image regions that share common visual properties -- offer a promising alternative to traditional attention mechanisms based on rectangular-shaped image patches in neural networks. Although previous work demonstrated that evolving a patch-based hard-attention module alongside a controller network could achieve state-of-the-art performance in visual reinforcement learning tasks, our approach leverages image segmentation to work with higher-level features. By operating on proto-objects rather than fixed patches, we significantly reduce the representational complexity: each image decomposes into fewer proto-objects than regular patches, and each proto-object can be efficiently encoded as a compact feature vector. This enables a substantially smaller self-attention module that processes richer semantic information. Our experiments demonstrate that this proto-object-based approach matches or exceeds the state-of-the-art performance of patch-based implementations with 62\% less parameters and 2.6 times less training time.
\end{abstract}



\maketitle

\section{Introduction}

\begin{figure}[t]
\includegraphics[width=0.45\textwidth]{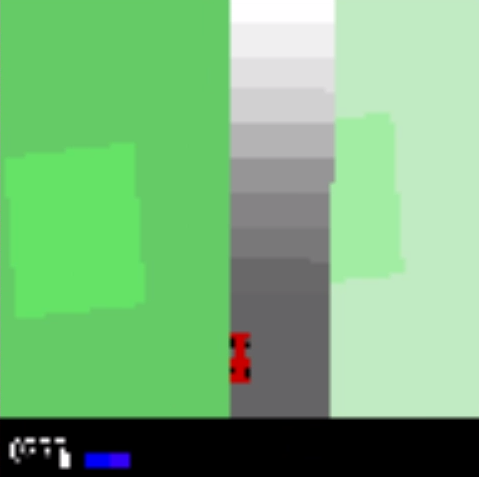}
\caption{Our attentional agent is capable of focusing on entire uniform regions instead of small fixed-size image patches.} \label{fig:region-attention}
\Description{}
\end{figure}

Visual attention mechanisms have emerged as a powerful solution for reducing computational complexity in high-dimensional perception tasks. By creating an information bottleneck between visual inputs and control networks, these mechanisms enable efficient processing of complex scenes \cite{recattention2014}. Recent work has demonstrated that evolving a hard-attention module jointly with an LSTM \cite{hochreiter1997long} controller can produce remarkably efficient agents that operate solely on small image patches \cite{attentionagent2020}. This approach not only yielded neural networks orders of magnitude smaller than competing methods but also achieved state-of-the-art results in challenging Reinforcement Learning environments like Car Racing and Doom Take Cover \cite{brockman2016openai}. The success stems from the attention layer's ability to filter irrelevant input regions, simplifying the controller's task while providing robust generalization and noise resistance.

We advance this line of research by replacing fixed-size, uniformly distributed patches with proto-objects -- coherent regions of locally uniform visual features \cite{Finkel:92} -- obtained through image segmentation \cite{fiorio1996two}. This shift in representation offers two key advantages. First, they provide a more compact representation, as most scenes decompose into fewer proto-objects than patches. Second, each proto-object encodes richer semantic information through a small descriptor vector capturing properties like shape, size, and color.

This proto-object approach enables a significantly streamlined architecture. The self-attention module becomes substantially smaller while processing higher-level features, leading to improved selection and better-filtered information for the controller, which can also be simplified. Our results in the Car Racing and Doom Take Cover environments \cite{brockman2016openai} demonstrate that this more efficient architecture matches or exceeds the performance of patch-based implementations while reducing the parameter count by 62\% with 2.6 times faster training.

\section{Background}

Modeling human visual attention has been an active research area over the past 35 years. Many different models of attention were proposed, which in addition to providing theoretical contributions to neuroscience and psychology, have demonstrated successful applications in computer vision and robotics \cite{borji2012state}. Early computational models focused primarily on bottom-up, saliency-based attention, while more recent approaches have incorporated top-down influences and object-based selection mechanisms.

The biological visual system provides crucial insights for designing efficient artificial vision systems. A fundamental constraint is that neural resources are limited -- \citet{KOCH20061428} demonstrated that retinal ganglion cells balance metabolic costs against information transmission, achieving highly efficient coding despite using relatively low firing rates. This suggests an evolutionary pressure toward strategic information bottlenecks rather than attempting to process all input equally. \citet{WALTHER20061395} showed that one such bottleneck occurs at the proto-object level, where coherent regions of the scene are selected for enhanced processing before full object recognition occurs. This allows the visual system to serialize complex scenes into manageable chunks while maintaining high coding efficiency.

\subsection{Types of Visual Attention Mechanisms}

Visual attention in biological systems operates through three primary mechanisms. Space-based attention operates on specific locations in the visual field, treating attention as a spotlight that enhances processing at selected spatial coordinates. Feature-based attention selectively enhances the processing of specific features (like color, orientation, or motion) across the entire visual field, regardless of spatial location. Object-based attention operates on perceptually grouped elements that form coherent objects, suggesting that attention selects entire object representations rather than just spatial locations or individual features \cite{Vecera1994DoesVA, Treue1999FeaturebasedAI, Chen2012ObjectbasedAA}.

\subsection{Self-Attention}

A key mechanism in modern computational attention is the self-attention layer. In its standard form \cite{vaswani2017attention}, self-attention operates on a set of $N$ input vectors, each of dimension $d_{in}$, linearly transforming them through learned weight matrices $\mathbf{W_Q}$ and $\mathbf{W_K}$ to obtain Query ($\mathbf{Q}$) and Key ($\mathbf{K}$) matrices:
\begin{equation}
\mathbf{S} = \text{softmax}(\frac{\mathbf{Q}\mathbf{K}^T}{\sqrt{d_k}})
\label{equ:attention}
\end{equation}
where $d_k$ is the dimension of the key vectors. 
The attention scores $\mathbf{S}$ show how related the input elements are. $\mathbf{S}$ is further combined with a matrix $\mathbf{V}$, which is also a linear transformation of the input, to form the contextual representation $\mathbf{A} = \mathbf{S}\mathbf{V}$, whose vectors contain the representation of each input considering the overall context. 
%The attention weights $\mathbf{A}$ determine how information from different input elements should be combined, by means of a matrix $\mathbf{V}$, which is also a linear transformation of the input.

\subsection{Proto-Objects and Information Bottlenecks}

Proto-objects are an intermediate representation between raw visual features and fully recognized objects \cite{WALTHER20061395, rensink2000dynamic}. They are formed during pre-attentive processing and represent coherent regions of the visual field that share common visual properties. These structures serve as potential candidates for attention before full object recognition occurs \cite{orabona2007proto}, allowing the visual system to efficiently prioritize processing resources.

Information bottlenecks in visual processing serve to compress the high-dimensional visual input into more manageable representations while preserving task-relevant information \cite{KOCH20061428, tishby2000information}. These bottlenecks can occur at various levels of processing, from early visual features to object recognition, and play a crucial role in managing the computational resources required for visual processing \cite{wolfe1994guided}. The formation of proto-objects itself represents a natural information bottleneck, as it reduces the complexity of the visual scene while maintaining behaviorally relevant information \cite{WALTHER20061395}.

\section{Related Work}

Our approach builds upon and extends several lines of research in visual attention, reinforcement learning, and deep learning. Here we review the most relevant prior work that informs and contextualizes our contribution, focusing particularly on approaches that address the challenges of efficient visual processing and attention mechanisms.

\subsection{Self-Interpretable Agents}

\begin{figure}[htb]
\includegraphics[width=0.45\textwidth]{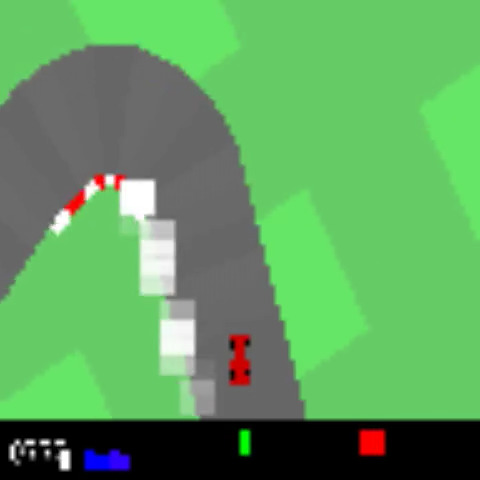}
\caption{The agent extracts patches from the current frame and selects the top-$k$ ones (highlighted in white) by a hard attention mechanism. By visualizing the selected patches, it is possible to have a better idea of the learned strategy, as well as making it easier for debugging. } \label{fig:attention}
\Description{}
\end{figure}

Recent work by \citet{attentionagent2020} explored the use of hard-attention mechanisms in reinforcement learning. Their approach divides the visual input into 7x7 patches and applies a modified self-attention mechanism to select only the top-$k$ patches for processing (see Fig. \ref{fig:attention}). This creates an information bottleneck that forces the model to focus on the most relevant inputs while ignoring others. When applied to vision-based reinforcement learning tasks, this approach enabled agents to achieve better performance, interpretability and computational efficiency with far fewer parameters than conventional approaches. Our method differs mainly by using proto-objects instead of patches, reducing the number of tokens and their dimensionality while adding higher-level information to them.

\subsection{Proto-Object Based Approaches}

\citet{orabona2007proto} demonstrated the effectiveness of proto-objects as the basic units of visual attention. Their work showed that using proto-objects as an intermediate representation naturally corresponds to potential objects in the scene, allowing for efficient processing while maintaining meaningful visual features.

The idea of using higher-level information from blobs as input was further explored by \citet{liang2015state} in their work on Blob-PROST (Pairwise Relative Offsets in Space and Time). Their approach demonstrated that visual features based on color blobs could achieve performance competitive with DQN \cite{mnih2013playingatarideepreinforcement} while using far fewer parameters. The core insight was representing game screens in terms of ``blobs'' - contiguous regions of same-color pixels that often correspond to game objects. This is a special case of our use of proto-objects, that are not restricted to same-color pixels. This representation was combined with position and relative offset features between blobs, capturing both spatial and temporal relationships. However, their representation consisted of more than 100 million binary features, while ours is based on compact representations of individual proto-objects that can be attented individually.

\subsection{Object-Based Reinforcement Learning}

While proto-object approaches focus on intermediate representations, some researchers have pushed further toward working with fully-fledged objects. \citet{Woof2018object} experimented with representations of high-level objects (e.g. identifiable entities in a game screen) rather than end-to-end video game playing. However, they assume the objects are previously given, which requires access to the game engine or specific object detectors. 

A more robust approach \cite{agnewunsupervised} demonstrated that operating directly on object-level representations can dramatically improve sample efficiency in reinforcement learning tasks. Their approach learns a succinct object representation from pixels without supervision by first oversegmenting images into primitive segments, modeling their dynamics, and then combining segments when they have similar dynamics.
Their method leverages the insight that important events tend to occur when objects interact, using this to guide exploration toward states involving object contacts. While their approach achieves impressive results -- learning up to 10,000 times faster than conventional deep RL methods and even outpacing human learning in some cases -- it requires significant computational overhead for object detection and tracking. This motivates the exploration of proto-object approaches, which provide a balance between processing efficiency and representational power.

\section{Proto-Object Attentional Agent}

Our work builds upon and connects several research directions in computer vision, deep learning, and evolutionary computation. We combine insights from biological models of visual attention, efficient neural architectures, and classical computer vision techniques to create a hybrid system that leverages the strengths of each approach. We apply hard-attention mechanisms to proto-objects rather than raw pixels or arbitrary patches. This approach implements an information bottleneck similar to what \citet{KOCH20061428} observed in biological systems, while operating on the semantically meaningful proto-objects described by \cite{rensink2000dynamic,orabona2007proto}. By selecting only the most relevant proto-objects for processing, we create an information bottleneck at a more semantically meaningful level than previous approaches.

This combination is particularly well-suited for neuroevolution, as the discrete selection of top-$k$ proto-objects and the transfer of coordinates to the controller create nondifferentiable operations that are challenging for gradient-based methods but natural for evolutionary approaches. Additionally, by forcing the model to be explicitly selective about which parts of the visual input it processes, we gain direct interpretability -- we can visualize exactly which proto-objects the model considers important for its decisions, providing insights into its decision-making process that are often lacking in traditional deep learning approaches.

\subsection{Implementation}

Our method consists of 5 main stages: convolution, quantization, segmentation, attention and control, described next. 

\subsubsection{Convolution}

The convolution stage aims to shift, rescale, filter and/or mix the original image channels, providing a pre-processed representation for the next stages. Particularly, in our experiments, we use a single convolutional layer with 3 1x1 filters. The choice of 3 filters is necessary for compatibility with the residual connection. It is possible to add more convolutional layers as long as they keep the same image size. In this case, adding a final layer with 3 filters is enough to bring down the number of dimensions to the same number of channels in the image. After that, we add the original image to the convolution output, forming a residual connection \cite{resnet2016}, whose function will become clear in the next stage.

\subsubsection{Quantization}

Quantization aims to reduce the amount of information to be processed in the next stages. In our experiments, we perform simple uniform quantization of the convolution output using 1 bit per channel (could be more for more complex tasks, and could even be evolved). As a result, we obtain an image with at most 8 distinct colors, each representing a different kind of segment. Note that, besides it being a simple fixed quantization, its combination with the convolutional layer before it results in an adaptive segmentation and quantization mechanism. 

This stage has synergy with the convolutions: the shift, rescale, and mix of the original image channels may put them into different quantization bins. But since there are discontinuous jumps in the evolution fitness surface necessary to find an appropriate segmentation, which can take some time for the evolutionary algorithm to figure out, we use the residual connection in the previous stage as a means to kickstart evolution from a trivial segmentation over the original image colors. Thus, the purpose of convolution is to change the segmentation away from the trivial one (if necessary).

\subsubsection{Segmentation}

Segmentation aims to create the proto-objects, i.e. descriptors for regions of semantically similar pixels received from the previous stages. In this work, we apply image labeling by color-connected regions \cite{fiorio1996two, silversmith_cc3d_2025}. From each extracted region, a set of attributes can be obtained, resulting in  $d_{in}$ features (regions 1 pixel wide or tall are treated as noise and ignored). 

After thorough experimentation, we ended up with a set of $d_{in}=11$ features, namely: quantized segment color (R, G, B), center of mass (X,Y), total area in pixels, bounding box width, bounding box height, bounding box area, aspect ratio and extent (bounding box area divided by region area). All of those can be easily and efficiently computed from the obtained regions, and will help the next stage to make more informed decisions. Orientation (correlation among pixel coordinates) could also be useful, but it added too much runtime overhead to our model and was left out. All values are normalized between -1 and  1, and aspect-ratio is also log-transformed such that 1 and -1 correspond to extreme ratios, while 0 means equal sides: 
$$NormAspectRatio = 2\frac{log(aspectRatio)}{log(max(imageWidth, imageHeight))} - 1$$

\subsubsection{Attention}
The attention module aims to model relations among the proto-objects identified in the segmentation stage.
The features for $N$ proto-objects are fed to our model's attention layer as a set of $N$ $d_{in}$-dimensional tokens, in the attention jargon \cite{vaswani2017attention}. The attention layer embeds these tokens into two $d_{q}$-dimensional vectors $\mathbf{Q}$ and $\mathbf{K}$. There is, however, an additional touch in our implementation: we add a Parametric Rectified Linear Unit (PReLU) layer before and after the linear transformations.  PReLU is a generalization of ReLU activation, where the slope of the negative part is adaptive ($PReLU(x) = max(ax, x)$) for each layer or neuron (the latter in our case). For only 15 additional parameters, this enables our attention layer to model more complex relations, as a single PReLU neuron was shown to solve the XOR problem \cite{pinto2024prelu}. In our case, it enables the selection of midrange values (like gray colors) when $a$ is negative (making the function nonmonotonic), which is not possible with pure linear layers. More layers of traditional self-attention could instead be used, but we opted for the simpler PReLU solution in this work to keep the number of parameters and runtime low.

Proceeding with the usual self-attention procedure, an attention matrix is computed by Eq. \ref{equ:attention}, and then an importance vector is obtained by row-wise summation.
Instead of the usual mixing of tokens with a $\mathbf{V}$ matrix performed in traditional self-attention, we simply perform top-$k$ proto-object selection over the resulting row-wise sum. 

We remark that the attention computation has quadratic asymptotic time complexity on the number of tokens $N$. Drastically reducing the number of tokens by using proto-objects instead of patches makes our attention module much faster.

In our experiments, we went to the extreme and set $k=1$ (coordinates of a single proto-object is passed to the controller, described next). This is possible because our attention module is more expressive than the original and proto-objects contain higher-level information, making a single well-selected proto-object enough for the controller to make its decisions (better selection means less work to the controller). It is also more biologically plausible, as we focus on a single visual item at time \cite{CARRASCO20111484}.

\subsubsection{Control}

Finally, the control stage selects an action to perform in the environment.
In our implementation, a transfer function $f(n)$ is applied to each feature vector from the selected proto-objects and the results are concatenated and fed as input to an LSTM \cite{hochreiter1997long} controller, which is responsible for learning temporal associations and producing the control output. 

In our case, $f(n)$ just returns the coordinates of the proto-object center of mass. More elaborate transfer functions could be used to feed the controller with more properties of each selected proto-object, but the center of mass was enough for our problems. This is possible because the joint evolution of the attention end control modules results in an implicit ``agreement``: by always selecting the same kind of proto-object (grass, track, etc...), there is no need for the controller to guess which is it. If attention focused on different kinds of proto-objects each time, it would not be possible to distinguish them solely by their coordinates, unless they consistently appeared on specific regions of the screen, being distinguishable by position (like the head-up display always at the bottom of the screen). They could also be distinguished by the controller  if the attention module consistently puts the same kinds of proto-objects into the same ranking spots (e.g., grass first, track second), but this is an additional piece of complexity to be learned. 

\subsection{Overview}

\begin{figure}[htb]
\centering
\includegraphics[width=0.47\textwidth, trim = 1cm 3cm 5cm 1cm, clip]{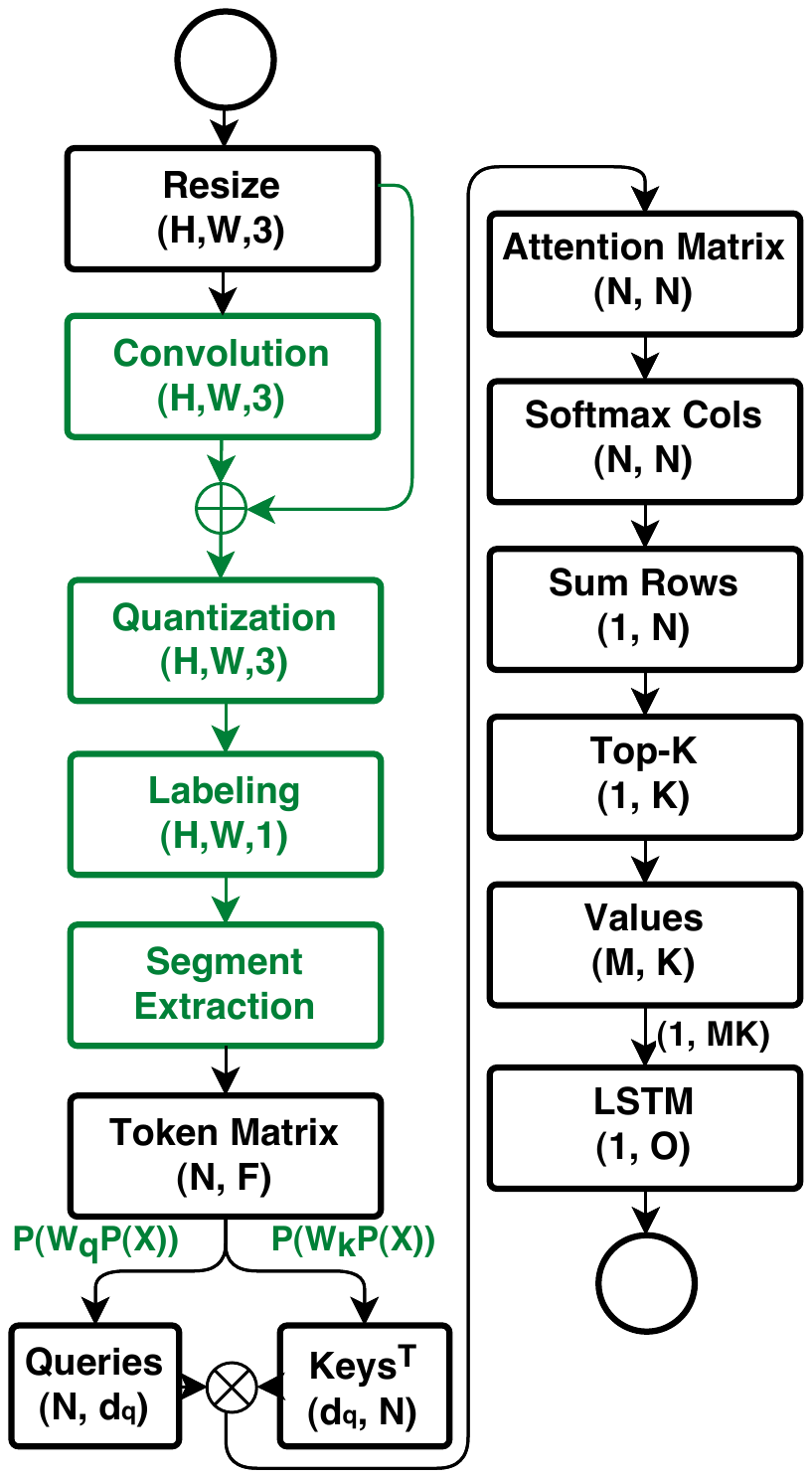}
\caption{Flowchart of our complete process. In our experiments, $H=W=96$, $F=11$ (number of segment features), $d_q=2$, $M=2$ (we use only the $x,y$ coordinates from each token), $k=1$ and $O=3$ (number of outputs for both environments). Any number of convolutional layers can be used, as long as they preserve image size and the last one has 3 filters to match the residual connection. We use one layer of 3 1x1 filters. Our quantization is set to 1 bit per channel (8 colors). $P$ is the $PReLU$ activation function. Green elements are new in relation to \cite{attentionagent2020}.} \label{fig:flowchart}
\Description{}
\end{figure}

A summary of the differences between our hyperparameter choices and the previous work based on image patches  \cite{attentionagent2020} is shown in Table \ref{tab:comparisonhyperparams}, as well as the resulting number of learnable parameters in each model, showing that our model is significantly (62\%) smaller in total, due to its compact attention layer and smaller bottleneck with $k=1$. The complete process can be seen in Fig. \ref{fig:flowchart}. Although this model is non-differentiable, it is learnable via derivative-free optimization methods such as CMA-ES \cite{hansen2006cma}.

\begin{table}[!htb]
  \caption{Comparison of hyperparameters and number of learnable parameters in patch-based model \cite{attentionagent2020} and proto-object-based model (ours). The latter uses 62\% less learnable parameters.}
  \label{tab:comparisonhyperparams}
  \begin{tabular}{ccc}
    \toprule
  & Patches \cite{attentionagent2020} & Proto-Objects (Ours) \\
    \midrule
      \multicolumn{3}{c}{\textbf{Model Hyperparameters}} \\
     \midrule
 Attention Input Size  ($d_{in}$) & 147 & 11 \\
 Embedding Size ($d$) & 4 & 2 \\
 K & 10 & 1 \\
 $f(n)$ Dimension & 2 & 2 \\
 LSTM Input Size & 20 & 2 \\
 \# of LSTM Neurons & 16 & 16 \\
     \midrule
      \multicolumn{3}{c}{\textbf{Number of Learnable Parameters}} \\
     \midrule
 Convolution & 0 & 12 \\ 
 Attention & 1184 & 63 \\ 
 LSTM & 2432 & 1280 \\
 Output & 51 & 51 \\
 \midrule
 \textbf{Total} & \textbf{3667} & \textbf{1406} \\
 \bottomrule
\end{tabular}
\end{table}

\section{Experiments and Results}

In order to compare our approach to the patch-based one of \cite{attentionagent2020}, we test it on the same environments of \cite{attentionagent2020}: CarRacing and DoomTakeCover \cite{brockman2016openai}. For both, we run CMA-ES with a population of 128 solutions for 1000 generations and evaluate models on 8 seeds every generation. Seeds are based on generation and repetition numbers. We test models every 100 generations on 400 new seeds and extract means and variances to produce 95\% confidence intervals. Statistical significance is obtained from two-sided Mann-Whitney U tests \cite{mannwhitney1947}. Note that the original experiments in \cite{attentionagent2020} ran for 2000 generations with 16 seeds each and a population of 256 solutions, so they are not directly comparable. We halved each of those hyperparameters due to hardware limitations, and performed the original experiments again in this new setup for fair comparison.

Our experiments ran on the following hardware setup: AMD Ryzen 5950X CPU, 128GB DDR4 3200 RAM and Nvidia RTX 3090 GPU. Training was paralellized over 32 threads, limiting each evaluation to a single thread. The patch-based solution took advantage of the GPU, but our method was optimized for CPU, as multi-label connected-components analysis and labeling did not fit well with the GPU.

\subsection{Car Racing}

This is a top-down racing environment with randomly generated tracks (as seen in Figs. \ref{fig:region-attention} and \ref{fig:attention}). It is visually simple enough to skip the convolution and quantization stages of our approach, but we perform them anyway in order to verify the generality of the method. The reward is $-0.1$ every frame, $-100$ for going far off-track (which also causes termination), and $+1000/N$ for every track tile visited, where $N$ is the total number of tiles visited in the track (tiles are visible as slightly distinct shades of gray), and it is considered solved above 900 points. This incentivizes the controller to be fast and accurate. There are 3 continuous actions:  steering (-1 is full left, +1 is full right), gas and braking. There is a version V2 of this environment available \footnote{https://gymnasium.farama.org/environments/box2d/car\_racing/}, but it uses Pygame \footnote{https://www.pygame.org}, which is slow. We use V0, which is twice as fast by using OpenGL, and implement our own optimizations that adds a further 2x speedup. There are no significant differences between both versions except for compatibility with the new API \cite{towers2024gymnasium} and better hardware and software compatibility.

%\begin{figure}[htb]
%\includegraphics[width=0.45\textwidth]{carracing.jpg}
%\caption{The Car Racing environment. It is visually simple %enough to not require the convolution and quantization steps, %but we apply them anyway to verify generality of our method. %Colored bars at the bottom, from left to right are speed %(white), ABS sensors (blue), steering wheel position (green) %and gyroscope (red).} \label{fig:carracing}
%\Description{}
%\end{figure}

\begin{figure}[htb]
\includegraphics[width=0.45\textwidth]{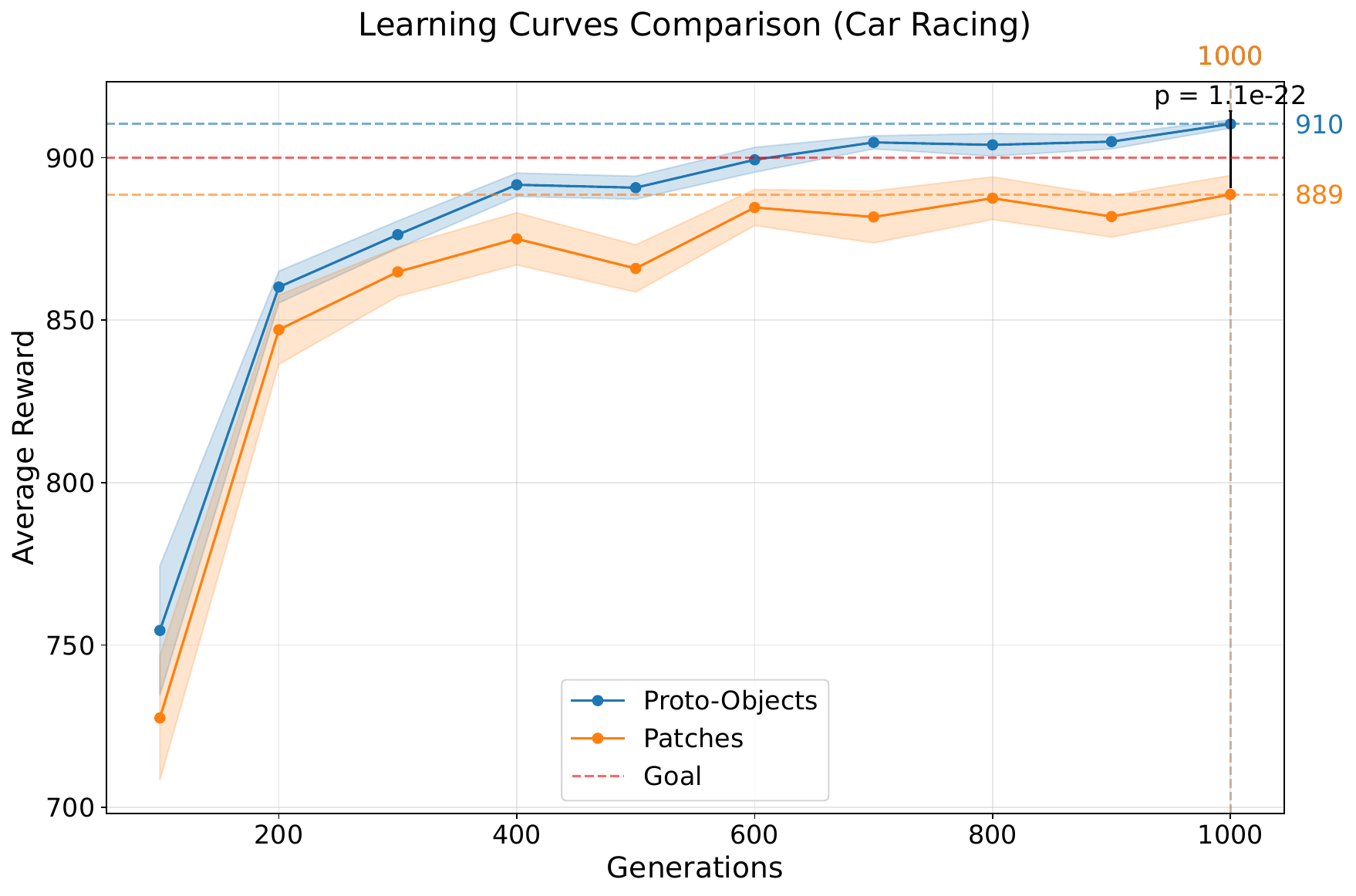}
\caption{Learning curve comparison in the Car Racing environment over 400 test runs out of training sample tracks for each shown generation. Our proto-object method achieves significantly better results after 1000 generations in relation to the patch-based of \cite{attentionagent2020}. Both peaked at 1000 generations.} \label{fig:lccar}
\Description{}
\end{figure}

Our method was more sample-efficient, with superior average score throughout the training, and achieved a significantly better ($p=1.1e{-22}$) score of $910.39$ after training (Fig. \ref{fig:lccar}). Moreover, as  Table \ref{tab:results} shows, it did so by using only $2\%$ the number of tokens per frame as the patch-based solution and $62\%$ less adjustable parameters. And despite running on CPU, it trained 2.7 times faster with respect to the patch-based method, which ran on GPU.

\begin{table}[htb]
  \caption{Comparison of results. The number of tokens per frame is fixed for the regular patch-based model, but variable for our proto-object-based model. Our method is currently optimized for CPU. In the number of tokens section, $n$ refers to number of frames, while in scores it equates to number of runs.}
  \label{tab:results}
  \begin{tabular}{ccc}
    \toprule
  & Patches \cite{attentionagent2020} & Proto-Objects (Ours) \\
    \midrule
      \multicolumn{3}{c}{\textbf{Number of Tokens and 95\% CI of Best Solution (n=800)}} \\
     \midrule
      Car Racing & 529 & 12.6 $\pm$ 0.26 \\
      Doom Take Cover & 529 & 10.7 $\pm$ 0.73 \\
     \midrule
      \multicolumn{3}{c}{\textbf{Best Score and 95\% CI After 1000 Iterations (n=400)}} \\
     \midrule
      Car Racing & 888.69 $\pm$ 5.84 & 910.39 $\pm$ 1.28 \\
      Doom Take Cover & 959.27 $\pm$ 58.85 &  930.68 $\pm$ 57.19 ($k=1$)\\ & & 1192.82 $\pm$ 75.26 ($k=10$) \\
     \midrule
      \multicolumn{3}{c}{\textbf{Training Time}}\\
     \midrule
      Car Racing & 97h (GPU) & 36.5h (CPU) \\
      Doom Take Cover & 85.5h (GPU) & 33h ($k=1$, CPU) \\ & & 55h ($k=10$, CPU)\\
  \bottomrule
\end{tabular}
\end{table}

\begin{figure*}[htbp]
\caption{Evolution of segmentation at 5 relevant points in time. The basic attention strategy (focus on the smaller grass region, highlighted in white with centroid in black; other centroids in pink) is learned early in the process, while segmentation keeps evolving until the end. Top-Left: Raw image after resizing and before segmentation. Top-Center (100 generations): The trivial segmentation from original colors is kept for the first couple hundred generations. Top-Right (300 generations): It learns to separate the track from the head-up display (HUD) below and differentiates some of the ABS sensors. Bottom-Left (800 generations): It swaps the track and HUD segmentations, making the car visible, differentiating the red corner markers and hiding the gyroscope indicators. It also differentiates the corner white markings from the score and speed indicator, and merges the ABS sensors again. Bottom-Center (900 generations): It differentiates the track segments. Bottom-Right (1000 generations): It gives up on the fine-grained segmentation of the track and goes back to the previous segmentation strategy, with very small and irrelevant changes in some pixels in the score.}
\includegraphics[width=0.32\textwidth]{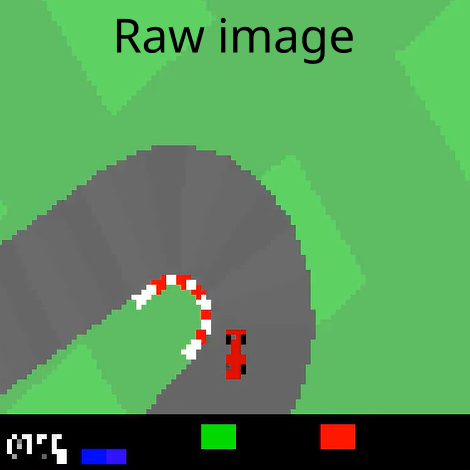}%
\includegraphics[width=0.32\textwidth]{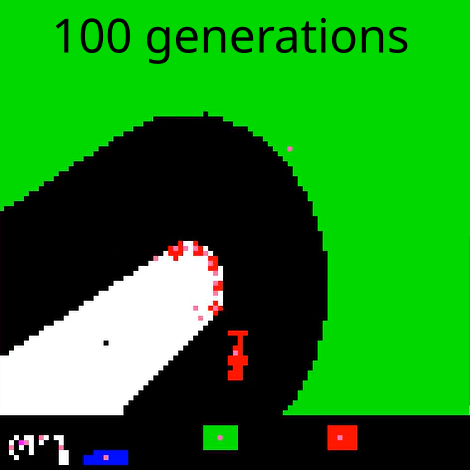}%
\includegraphics[width=0.32\textwidth]{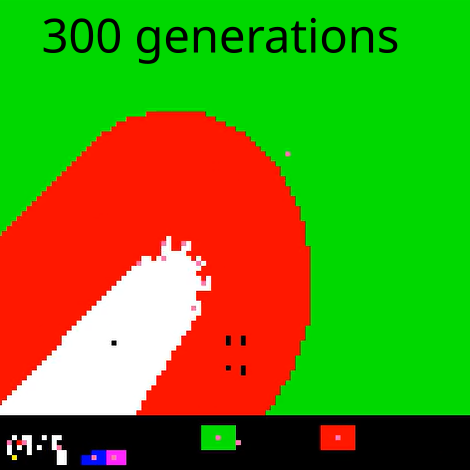}\\%
\includegraphics[width=0.32\textwidth]{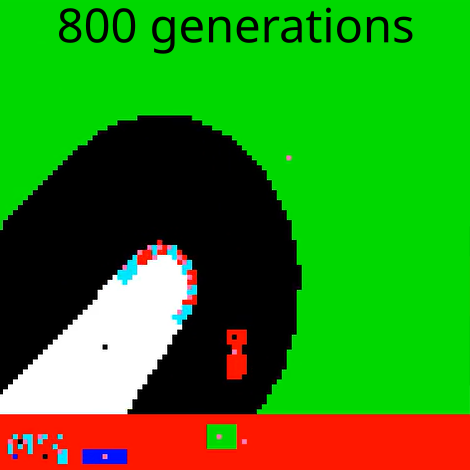}%
\includegraphics[width=0.32\textwidth]{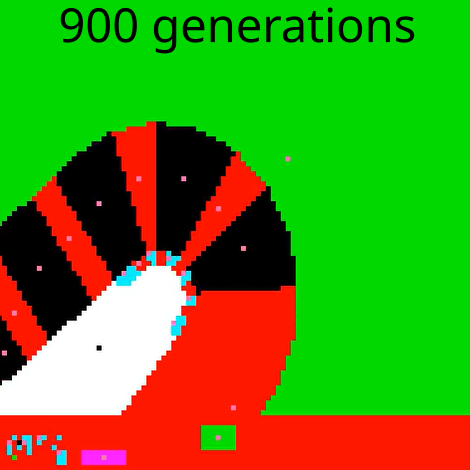}%
\includegraphics[width=0.32\textwidth]{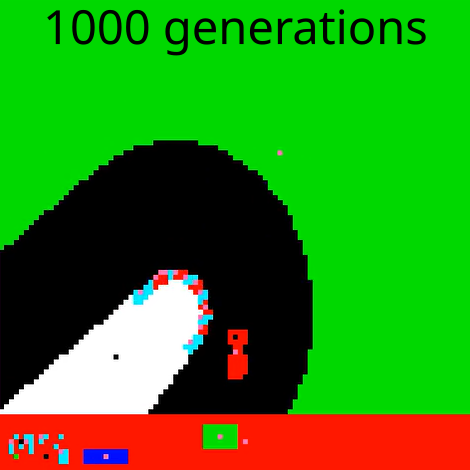}%
\label{fig:attevo}
\Description{}
\end{figure*}

An interesting aspect of this experiment is to observe the evolution of segmentation and attention, as Fig \ref{fig:attevo} shows. The solution starts with the trivial quantization over the task's original colors, but since the track is dark, it gets merged with the black head-up display (HUD) at the bottom of the screen. Nevertheless, it already knows how to focus on the smaller grass region, as it usually points at the direction the car must turn. At 300 generations it learns to separate the track from the HUD, while at 800 generations it separates the car and the red corner markings from the track. While the car is useless (it is always at the same place and was even merged with the track in other experiments), the red markings can reinforce the correct turning side by ``voting'' (as queries) on their adjacent grass region. At 900 generations, it learns to segment the track tiles, but discards it in the final solution. %This entire sequence can be seen in Fig. \ref{fig:attevo}.

\subsection{Doom Take Cover}

\begin{figure}[htb]
\includegraphics[width=0.45\textwidth, trim = 0cm 1.35cm 0cm 0cm, clip]{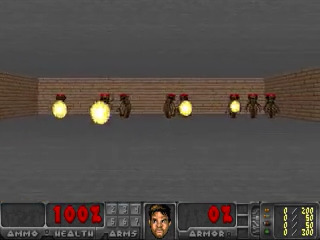}
\caption{The Doom Take Cover environment. The agent has to avoid the fireballs shot by the enemies in the back of the room by moving left and right. The higher number of colors requires our convolution and quantization components to prevent a huge number of segments.} \label{fig:takecover}
\Description{}
\end{figure}

This task is based on the game Doom, which is visually more complex and features much more colors than the previous task (see Fig. \ref{fig:takecover}), making the convolution and quantization steps strictly necessary to prevent a huge number of segments. It takes place in a rectangular room. The agent is spawned along the wall, and monsters are constantly and randomly spawned along the opposite wall. They keep shooting fireballs at the agent, which must avoid them to survive. The agent gets 1 reward point for every tic  alive and has 3 discrete actions: move left, right or stand still.

\begin{figure}[htb]
\includegraphics[width=0.45\textwidth]{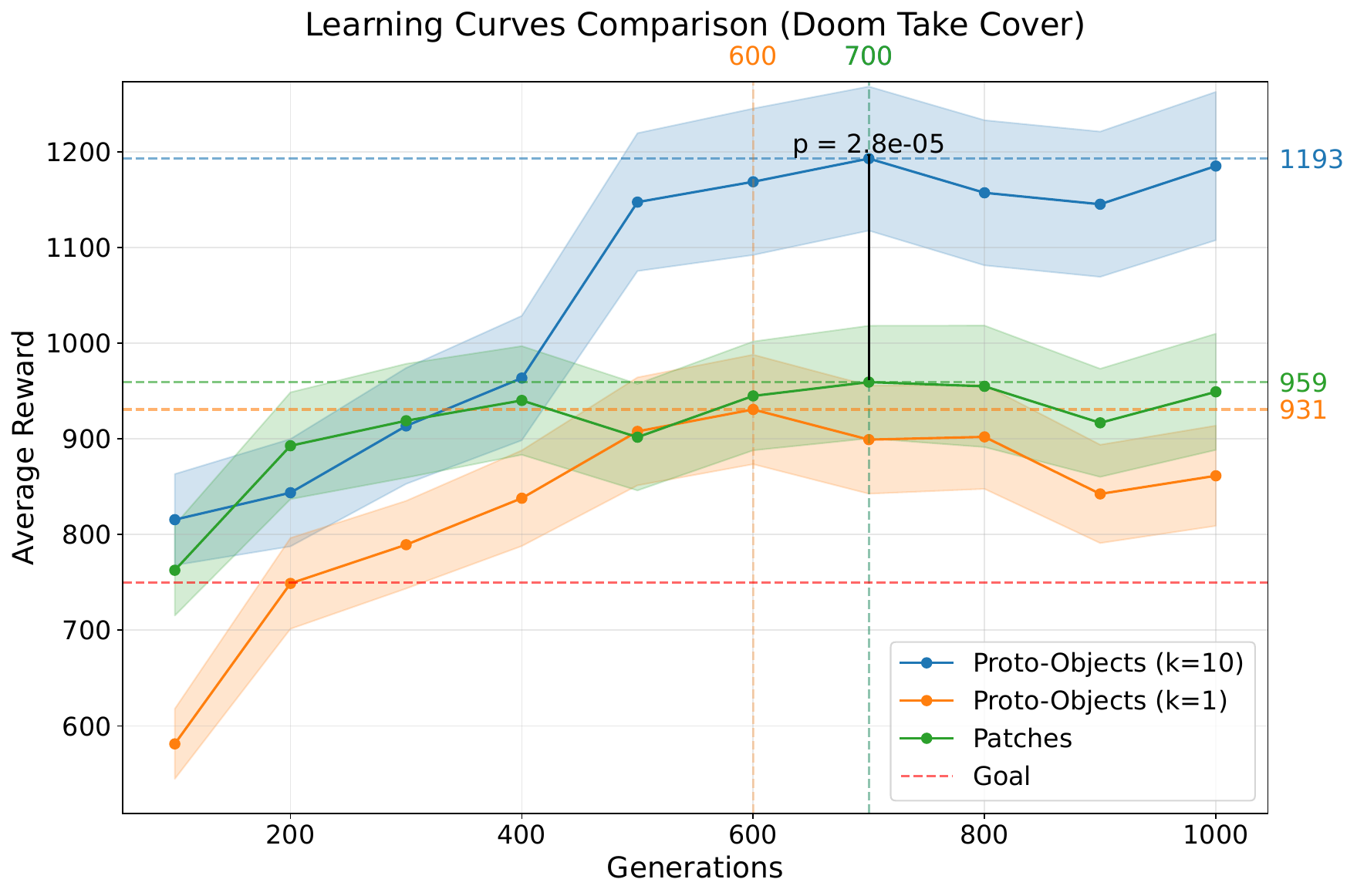}
\caption{Learning curve comparison in the Doom Take Cover environment over 400 test runs out of training sample seeds for each shown generation. Our method with $k=1$ showed lower sample-efficiency but achieved similar performance after 1000 generations in relation to \cite{attentionagent2020}, which achieved peak performance at 700 generations ($p=0.414$ between best solutions). However, with $k=10$ (same as the patch-based setup) and further adjustments detailed in the text, our solution had better sample-efficiency and achieved significantly higher ($p=2.8e{-5}$) performance (1193 score at 700 generations.} \label{fig:lcdoom}
\Description{}
\end{figure}

Learning curves are shown in Fig. \ref{fig:lcdoom}. We observe that our approach had slightly lower sample-efficiency, needing more generations to match the patch-based model performance ($p=0.414$). We also experimented with $d_{q}=4$, $k=10$ (same as the patch-based setup) and 3x3 convolutions (2671 parameters) and this solution had better sample-efficiency and achieved significantly ($p=2.8e{-5}$) higher  performance at a 1193 score in 55h of training. We hypothesize that the performance degradation was due to $k=1$, meaning that the LSTM is under much harder work to keep up with multiple interest proto-objects on screen, or even missing some of them entirely, while also learning to discard the wall proto-objects that activate when there are no projectiles on screen.

Table \ref{tab:results} also shows that the number of extracted proto-objects was low for this environment as well, demonstrating that our preprocessing steps are effective in reducing and uniformizing the visual complexity of different domains, while keeping necessary information for decision making. The training time was 2.6 times faster for $k=1$ and 1.6 times faster for $k=10$.

\begin{figure}[htb]
\includegraphics[width=0.45\textwidth]{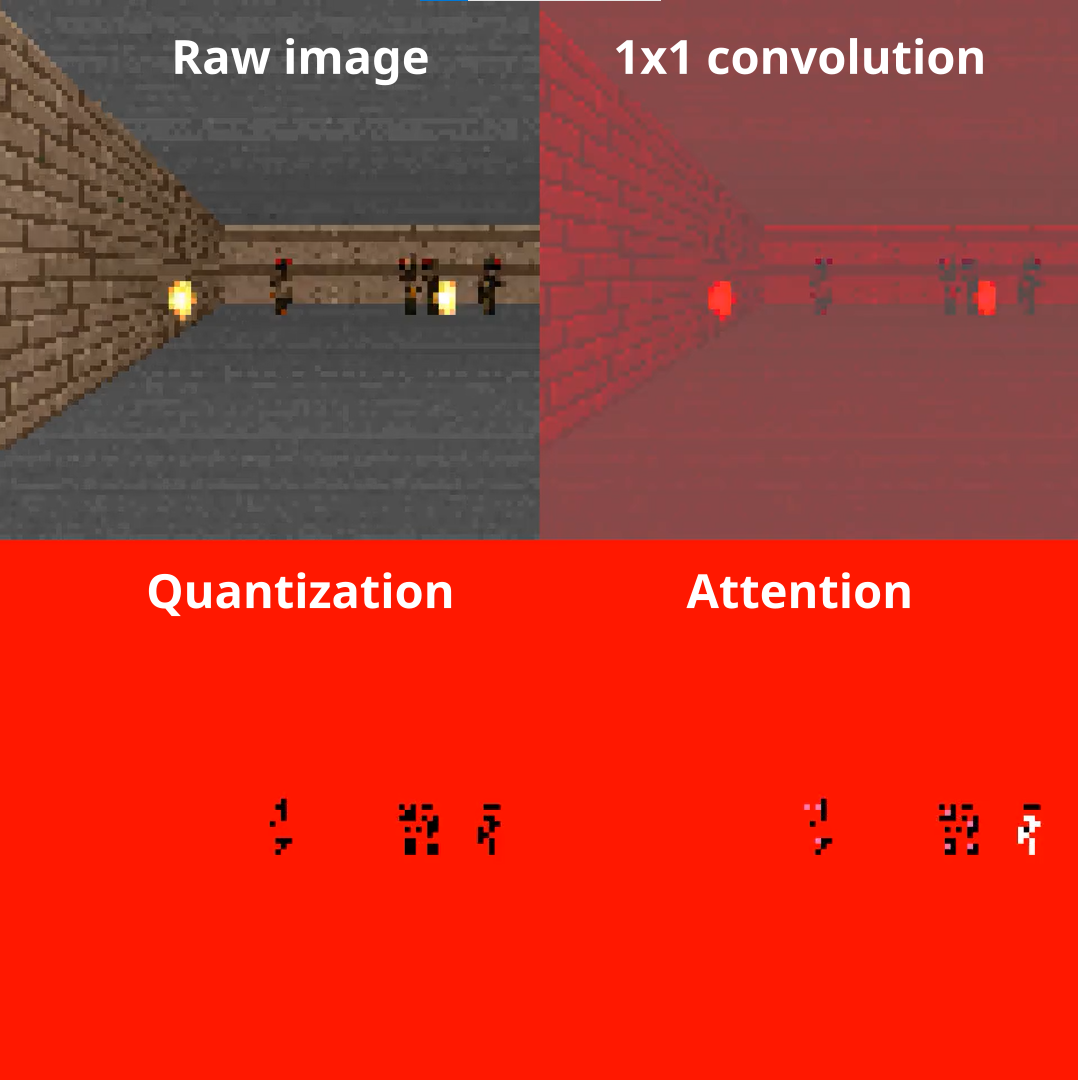}
\caption{Processing stages in the Doom Take Cover environment. Top-left: raw image resized to 96x96. Top-right: 1x1 convolution + residual. Bottom-left: quantization. Bottom-right: attention. Note that this solution completely ignores the fireballs since the quantization stage.} \label{fig:doom}
\Description{}
\end{figure}

The key processing stages in the Doom environment are illustrated in Fig. \ref{fig:doom}: image resizing, 1x1 convolution, color quantization, and attention ($k=1$). Remarkably, the evolved agent adopts a surprisingly minimalist strategy, ignoring seemingly critical elements like incoming fireballs. Instead, it focuses exclusively on the rightmost monster on screen while executing a rhythmic left-to-right movement pattern. This strategy matches the performance of the patch-based model, despite the latter attending to both fireballs and walls. The equivalence to our simple approach suggests that the patch-based model's LSTM might also be relying primarily on periodic movement and ignoring projectile coordinates. This strategy proves effective because the monsters' projectiles target the agent's current position -- continuous movement therefore serves as a robust avoidance technique, regardless of the specific locations of incoming fire. However, our $k=10$ agent seems more reactive to fireballs.

\section{Conclusion and Future Works}

We have presented a novel representation for bottleneck-attention-based agents in visual tasks that operates on proto-objects rather than raw pixels or image patches. By working with these pre-attentional primitive objects, obtained through classical computer vision methods, we achieved comparable or superior performance while dramatically reducing the number of tokens to attend and their dimensionality, as well as training time compared to previous solutions. The success of this hybrid approach highlights one of the key advantages of evolutionary methods for training such models: the freedom to combine differentiable and non-differentiable components without being constrained by the requirements of gradient-based optimization. Nevertheless, developing a fully differentiable version of our solution remains an attractive direction for future work, as it could substantially improve sample efficiency.

Our experiments revealed that bottleneck attentional models are susceptible to local maxima during evolution. The dual-module architecture (attention and control) makes it challenging to discover new attention strategies once an approach becomes established, as the controller adapts specifically to the current attention mechanism. Any significant changes to the attention module risk disrupting this delicate balance. We hypothesize that CMA-ES may be too greedy for this architecture, and alternatives like differential evolution \cite{Storn1997DifferentialE} might be more suitable by allowing multiple attention strategies to evolve in parallel.

We demonstrated that by enhancing the attention layer, sending coordinates of a single proto-object to the controller is sufficient to produce effective policies. This works because the LSTM can maintain and update an internal state representation across frames, deciding what information to preserve or discard. This approach aligns well with biological eye movements, where focus necessarily shifts between individual locations or objects \cite{CARRASCO20111484}. However, this simplified information flow comes at the cost of longer learning times when there are multiple relevant entities on screen, as the attention module must attend to them all and the controller must develop sophisticated memory management strategies. One potential solution is to decouple memory and control, possibly by implementing attention mechanisms over recently attended coordinates to generate fixed-size embeddings \cite{reimers2019sentencebertsentenceembeddingsusing} for the controller. This could be further extended to include adaptive storage and retrieval from vector databases.

Several promising directions for future research emerge from this work. Feedback signals from the controller could modulate attention, enabling active top-down strategies. This would require enriching the information flow from the attention module to help the controller interpret incoming signals. Multi-head attention represents another natural extension. The approach could potentially scale to full object recognition by incorporating additional convolutional layers and processing depth (when available) and motion information. Self-attention mechanisms might enable autonomous grouping of regions into higher-level entities, while cross-attention could facilitate object tracking across frames.

Finally, a crucial next step is validating our approach on real-world images and determining whether increased complexity in the convolution and quantization stages is necessary, or if patch-based approaches prove more effective in such scenarios. Success in this domain could lead to more efficient robot and self-driving car systems, reducing computational requirements while enabling more sophisticated intelligence per processing unit.

\begin{acks}
The authors would like to acknowledge  FAPERGS (Notice 10/2021 – ARD/ARC) for the financial support. This study was also supported by the Federal Institute of Education, Science and Technology of Rio Grande do Sul (IFRS).
\end{acks}

\bibliographystyle{ACM-Reference-Format}
\bibliography{arxiv}

\end{document}